\documentclass[conference]{IEEEtran}
\usepackage[numbers,sort&compress]{natbib}
\usepackage{graphicx}
\usepackage{dirtytalk}
\usepackage{etoolbox}
\usepackage{booktabs}
\usepackage{multirow}

\usepackage[table,xcdraw]{xcolor}
\makeatletter

\patchcmd{\@makecaption}
 {\\}
 {.\ }
 {}
 {}
 
\ifCLASSINFOpdf
\else
\fi
\usepackage{amsmath}
\usepackage{amssymb}
\usepackage{threeparttable}
\usepackage{scalerel}
\usepackage{tikz}
\usetikzlibrary{svg.path}

\definecolor{orcidlogocol}{HTML}{A6CE39}
\tikzset{
    orcidlogo/.pic={
        \fill[orcidlogocol] svg{M256,128c0,70.7-57.3,128-128,128C57.3,256,0,198.7,0,128C0,57.3,57.3,0,128,0C198.7,0,256,57.3,256,128z};
        \fill[white] svg{M86.3,186.2H70.9V79.1h15.4v48.4V186.2z}
        svg{M108.9,79.1h41.6c39.6,0,57,28.3,57,53.6c0,27.5-21.5,53.6-56.8,53.6h-41.8V79.1z M124.3,172.4h24.5c34.9,0,42.9-26.5,42.9-39.7c0-21.5-13.7-39.7-43.7-39.7h-23.7V172.4z}
        svg{M88.7,56.8c0,5.5-4.5,10.1-10.1,10.1c-5.6,0-10.1-4.6-10.1-10.1c0-5.6,4.5-10.1,10.1-10.1C84.2,46.7,88.7,51.3,88.7,56.8z};
    }
}

\newcommand\orcidicon[1]{\href{https://orcid.org/#1}{\mbox{\scalerel*{
                \begin{tikzpicture}[yscale=-1,transform shape]
                \pic{orcidlogo};
                \end{tikzpicture}
            }{|}}}}

\newcommand{\figref}[1]{Fig.\hspace{1mm}\ref{#1}}
\newcommand{\tabref}[1]{Table\hspace{1mm}\ref{#1}}

\ifCLASSOPTIONcompsoc
\usepackage[caption=false,font=normalsize,labelfont=sf,textfont=sf]{subfig}
\else
\usepackage[caption=false,font=footnotesize]{subfig}
\fi
\usepackage{hyperref}

\begin{document}

\title{LeafGAN: An Effective Data Augmentation Method for Practical Plant Disease Diagnosis}

\author{\IEEEauthorblockN{Quan Huu Cap\IEEEauthorrefmark{1}, Hiroyuki Uga\IEEEauthorrefmark{2}, Satoshi Kagiwada\IEEEauthorrefmark{3}, and Hitoshi Iyatomi\IEEEauthorrefmark{1}}
\IEEEauthorblockA{huu.quan.cap.78@stu.hosei.ac.jp\quad uga.hiroyuki@pref.saitama.lg.jp\quad kagiwada@hosei.ac.jp\quad iyatomi@hosei.ac.jp}
\IEEEauthorblockA{\IEEEauthorrefmark{1}Applied Informatics, Graduate School of Science and Engineering, Hosei University, Tokyo, Japan}
\IEEEauthorblockA{\IEEEauthorrefmark{2}Saitama Agricultural Technology Research Center, Saitama, Japan}
\IEEEauthorblockA{\IEEEauthorrefmark{3}Clinical Plant Science, Faculty of Bioscience and Applied Chemistry, Hosei University, Tokyo, Japan}}

\maketitle
\begin{abstract}
    Many applications for the automated diagnosis of plant disease have been developed based on the success of deep learning techniques. 
However, these applications often suffer from overfitting, and the diagnostic performance is drastically decreased when used on test datasets from new environments. 
In this paper, we propose LeafGAN, a novel image-to-image translation system with own attention mechanism. 
LeafGAN generates a wide variety of diseased images via transformation from healthy images, as a data augmentation tool for improving the performance of plant disease diagnosis. 
Thanks to its own attention mechanism, our model can transform only relevant areas from images with a variety of backgrounds, thus enriching the versatility of the training images. 
Experiments with five-class cucumber disease classification show that data augmentation with vanilla CycleGAN cannot help to improve the generalization, i.e., disease diagnostic performance increased by only 0.7\% from the baseline. 
In contrast, LeafGAN boosted the diagnostic performance by 7.4\%. 
We also visually confirmed the generated images by our LeafGAN were much better quality and more convincing than those generated by vanilla CycleGAN. 
The code is available publicly at: \url{https://github.com/IyatomiLab/LeafGAN}.
\end{abstract}

\begin{IEEEkeywords}
image-to-image translation, plant disease diagnosis, data augmentation, generative adversarial network.
\end{IEEEkeywords}

\section{Introduction}
\begin{figure}[!t]
\centering
\includegraphics[width=0.99\linewidth]{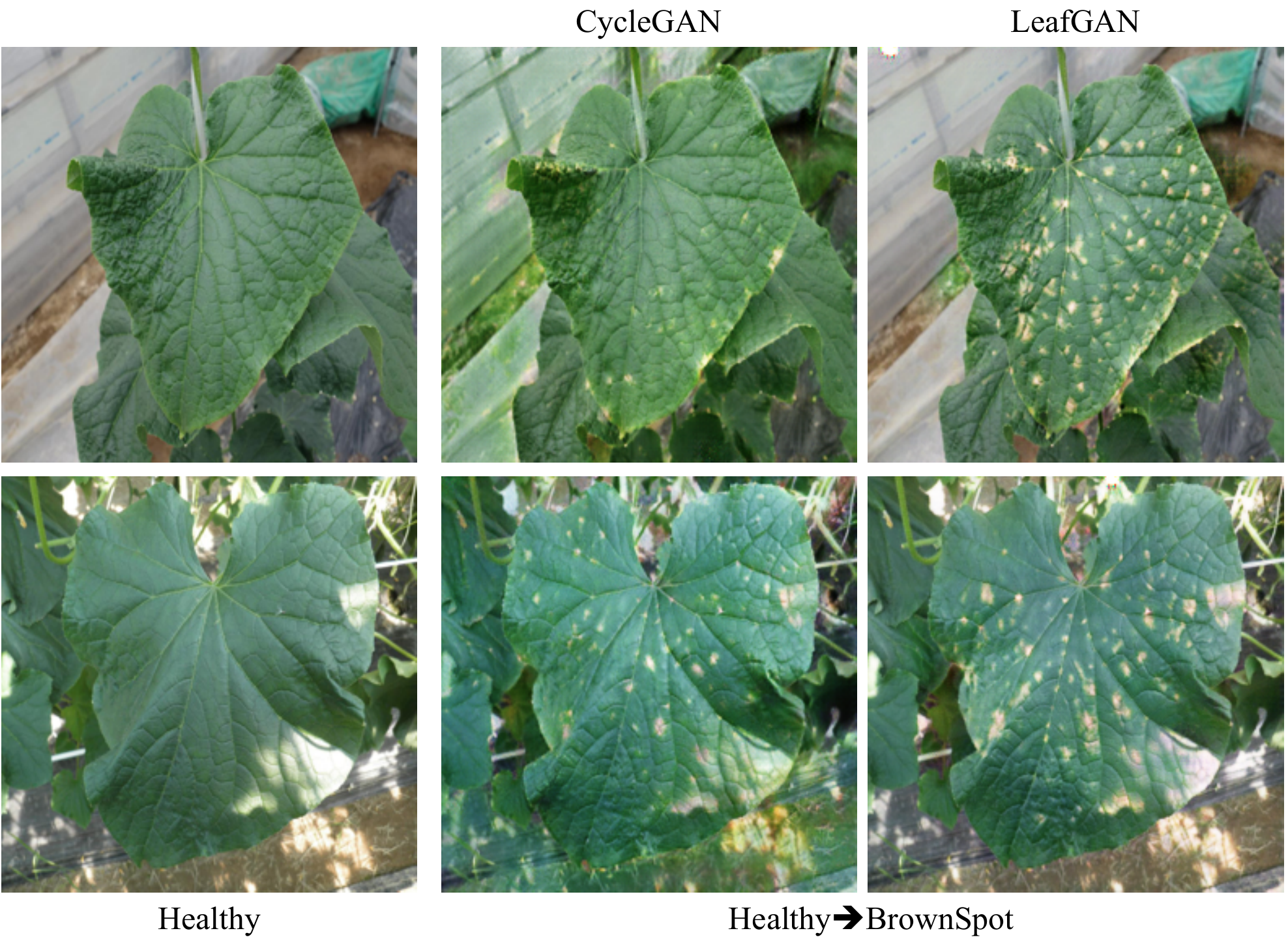}
\caption{
    Comparison of the original CycleGAN and our LeafGAN to transform images of healthy plants to diseased ones (brown spot in this case). The CycleGAN model transforms not only the leaf regions but also the background, and as a result, the generated images have an unrealistic quality compared to the proposed LeafGAN method.
}
\label{fig:fig_1}
\end{figure}
\IEEEPARstart{T}{he} automated diagnosis of plant disease is one of the most active research fields in agriculture. 
Detecting plant diseases in a timely and accurate manner is essential in ensuring global food security and the sustainability of agroecosystems \cite{strange2005plant, savary2019global}. 
In recent years, deep learning has revolutionized the field of computer vision, and is now becoming a standard tool for many applications. 
Many deep learning-based techniques for the automated diagnosis of plant disease have been developed with the aim of supporting farmers and reducing losses in terms of plant productivity. 

Mohanty \emph{et al.} \cite{mohanty2016} analyzed a total of 54,306 plant leaf images consisting of 38 crop-disease pairs (including 26 plant diseases) using the PlantVillage dataset \cite{hughes2015open}. 
They evaluated their system with cross-validation and reported a mean accuracy of 99.3\%. 
Using this large-scale open dataset, diagnostic studies of apples \cite{liu2018identification, wang2017automatic}, tomatoes \cite{durmucs2017disease, atabay2017deep}, and plural targets \cite{ferentinos2018deep} also showed mean accuracies of more than 90\%. 
Although the methods described above attained excellent results, Mohanty \emph{et al.} reported that the performance of their approach was reduced to just above 31\% when tested using a set of on-site plant images \cite{mohanty2016}. 
This was due to fact that the images in these datasets were taken  in a laboratory setting (i.e., each leaf was cropped in advance and the images had a uniform background); hence, effects from other factors such as the background and variation in photographic conditions had serious impacts on performance. 
Ferentinos \cite{ferentinos2018deep} also pointed out that when the model was trained solely on images taken under laboratory conditions and tested on images taken under cultivation conditions, success rates dropped significantly from 99.5\% to about 33\%. 
We have high confidence that other related studies will show similar low diagnostic performance under practical conditions. 

Many plant disease diagnosis systems for real cultivation conditions have also been proposed. 
In pioneering work, Kawasaki \emph{et al.} \cite{kawasaki2015} trained a three-layer convolutional neural network (CNN) to diagnose three classes of cucumber diseases (two classes of diseased and one of healthy) on images from a real farm, in which the target objects appear with complex backgrounds. 
Their model achieved an average accuracy of 94.9\%. 
Similar studies on cucumber \cite{fujita2016, hiroki2018diagnosis} have also conducted. 
Fuentes et al. \cite{fuentes2017robust} combined three object detection methods to perform disease detection and diagnosis simultaneously on wide-range tomato dataset, achieving 86.0\% mean average precision. 
Ramcharan et al. \cite{ramcharan2017deep} reported an overall 93\% classification accuracy on on-site six cassava leaf classes. 
DeChant et al. \cite{dechant2017automated} proposed an automated system to identify northern leaf blight lesions on field-acquired images of corn plants and achieve 96.7\% accuracy on test set. 
Using a dataset of natural rice images, Lu et al. \cite{lu2017identification} trained their system to identify 10 common rice diseases. 
Under the 10-fold cross-validation strategy, the proposed CNN model achieves an accuracy of 95.48\%. 
A systematic review of these techniques can be found in \cite{brahimi2018deep, singh2018deep, boulent2019convolutional}. 

Despite the success of the above methods, several essential problems still remain. 
\emph{Firstly}, deep learning-based systems need a huge number of training images. 
Unlike other general computer vision tasks, labeling disease datasets requires solid biological knowledge. 
Moreover, in order to collect gold standard datasets of diseases, the plants must be grown in a strictly controlled and isolated environment to avoid contamination, which is generally labor-intensive and very expensive. 
\emph{Secondly}, practical plant disease datasets are often imbalanced. 
Although the target plants are grown in a tightly controlled environment as described above, disease development is also strongly influenced by ambient conditions such as weather, temperature and vector-borne insects. 
Therefore, several diseases are difficult to collect, and the obtained datasets often have imbalanced amount on each class. 
Although several techniques have been proposed to address this data imbalance problem \cite{lin2017focal,cui2019class}, disease classification models are generally biased toward classes with more samples and higher variation \cite{fuentes2018high}. 
\emph{Thirdly}, the overfitting problem is particularly serious in plant diagnosis tasks, since the image features that provide diagnostic clues (i.e., evidence for classification) are typically much smaller than in general object recognition problems. 
Particularly in early-stage cases, the clues for diagnosis may consist only of a tiny dot or faint wrinkles in the image. 
This is the main problem that is going to be addressed in this paper. 
Image-based plant diagnosis is a particularly difficult task due to the fine-grained object recognition required. 
In general, a deep classifier such as a CNN tends to capture the image characteristics (brightness, color) of a large area, rather than a faint feature that may indicate disease. 
In addition, when evaluating a classifier using a dataset divided into training, validation, and test sets (where cross-validation is applied), the \say{latent similarity} within the dataset (such as the background, brightness and/or distance between target and camera etc.) works as a positive bias, and generally improves only the superficial diagnostic accuracy, while the accuracy when evaluated on other unknown environments becomes very low \cite{mohanty2016, ferentinos2018deep, quan2018, saikawa2019aop, suwa2019comparable}. 
For example, in the cucumber disease diagnosis from wide-angle images, the diagnostic performance on the same farm showed 86.0\% in F1-score, but it dropped to 20.7\% on a different farm \cite{quan2018}. 
Other evidence conﬁrming the overﬁtting of models in plant diagnosis tasks has been shown in our previous studies \cite{saikawa2019aop, fujita2018practical} by using Grad-CAM \cite{selvaraju2017grad} to visualize the key regions of diagnostic evidence. 
Although these models provided a high diagnosis accuracy of over 90\% on this dataset, the backgrounds were sometimes considered as diagnostic regions. 

The most plausible reason for this is that when collecting a dataset, the foreground objects in each image class tend to be incidentally correlated with similar backgrounds. 
A lack of background diversity could be a distractor, meaning that the model sometimes responds to the background rather than discriminative targets (i.e., leaf regions). 
One possible solution for this is to remove the background from the region of interest (RoI) as in our proposal anti-overﬁtting pretreatment (AOP) network \cite{saikawa2019aop}. 
The network segments the leaf areas before training disease classiﬁers, in order to reduce the negative impact of the background in terms of causing overﬁtting. 
We conﬁrmed that our AOP signiﬁcantly improved the classiﬁcation performance in a practical setting. 
However, this approach requires a large amount of expensive masking data and may eliminate surrounding information that is important for diagnosis (e.g., the lighting conditions of the picture, indicators of infection). 
Furthermore, we believe that the “latent similarity” within the dataset such as brightness, lighting, and/or distance between target and camera etc. still remains even on the segmented images and could cause difficulties for diagnosing on unseen data. 

In general, the background diversity of disease images tends to be limited, especially when plants are grown in a controlled environment to ensure the quality of training labels. 
However, collecting healthy images is relatively easy. 
In these situations, we can assume that if we could transform the wide variety of healthy images (including backgrounds) into disease cases, we could build a more divergent and reliable disease dataset. 
As a result, we expect to both improve the performance of diagnosis and to reduce the cost of labeling. 

Recently, an excellent image-to-image translation method called CycleGAN \cite{zhu2017unpaired} has been shown to have outstanding performance and has become a standard method of generating appealing images. 
CycleGAN removes the need for paired label training data by introducing the cycle-consistency loss, based on the assumption that the image generated from the source domain should be able to be transformed back to its original form. 

Based on the superiority of CycleGAN, several methods have been developed for application in the field of plant science. 
Tian \emph{et al.} \cite{tian2019detection} applied CycleGAN as a data augmentation method to generate more data on diseased apples to train their apple lesion detection system. 
However, since CycleGAN generates images that are close to the distribution of the original training data, the effect of adding these generated images to the training set was limited. 
In addition, because the original CycleGAN itself has no explicit attention mechanism, it tends to transform the entire image from the source to the target domain, rather than transforming the specific objects (i.e., the apple in this case). 
As a result, a significant number of the generated images are of low quality. 

Nazki \emph{et al.} \cite{nazki2020unsupervised} improved CycleGAN by introducing an additional perceptual loss \cite{johnson2016perceptual} in order to generate more natural images. 
Their model so-called AR-GAN transformed healthy tomato leaves into six different kinds of disease, and they claimed that their proposal could signiﬁcantly improve disease classiﬁcation performance compared to other classical data augmentation techniques. 
However, AR-GAN was trained on tomato images which have no complex backgrounds (i.e., almost entire area of each image is tomato leaves) and based on our preliminary experiments, it mostly failed to transform the symptoms on the images which include practical backgrounds like ours. 
Moreover, their disease classiﬁer was tested on a dataset that was split from the same population as the training dataset, the results must be biased due to the “latent similarity” among the datasets as mentioned earlier. 
Therefore, no essential results have been conﬁrmed. 

In order to overcome these limitations and achieve a practical method of image augmentation, we propose an image-to-image translation system named LeafGAN for generating images of leaves from diseased plants. 
LeafGAN determines the area of the image that is relevant for diagnosis, and translates only that area from the source to the target domain. 
The key idea is to develop a segmentation module that segments the area of interest (i.e., the leaf region) from the background, and which can help in guiding our LeafGAN model to pay attention to the RoIs. 
Similar to our study, there have been studies to improve CycleGAN by introducing the attention mechanism \cite{mejjati2018unsupervised, chen2018attention, yang2019show}. 
All of those studies added an attention network to each generator in CycleGAN and produce attention maps to guide the generator transforming the most discriminative regions between the source and target domains only. 
The attention networks are then trained simultaneously with CycleGAN model. 
Different from their works where those attention networks are sensitive to initialization and require careful care in training, our segmentation module can be trained very quickly, and easily to achieve effective segmentation results. 
Moreover, our segmentation module is trained separately, and we use only one segmentation network for both generators in our LeafGAN. 

We observe that LeafGAN not only generates high-quality images compared to CycleGAN, but also makes disease diagnosis systems more robust against unseen data by adding these generated images as training resources. 
Our contributions can be summarized as follows:
\begin{itemize}
    \item We propose the LeafGAN model for practical plant disease diagnosis. This is an effective and easy-to-implement data augmentation tool that generates natural, high-quality disease images from healthy images while preserving a wide variety of backgrounds. 
    \item We demonstrate the effectiveness of LeafGAN in terms of improving the generalizability of diagnostic systems. 
    Training with the augmented data generated by our system improves the average diagnostic performance by 7.4\% on different unseen images taken from other farms while generated images from CycleGAN only help improve by 0.7\%. 
    \item As a key module of LeafGAN, we introduce a novel label-free leaf segmentation module called LFLSeg, composed of a weakly supervised segmentation network that learns how to segment the leaf region without the need of expensive masking data. 
    LFLSeg provides guidance during training that helps the network to focus attention on the leaf regions for image-to-image translation in LeafGAN.
\end{itemize}

\section{Proposed method - LeafGAN}
\begin{figure}[!t]
\centering
\includegraphics[width=0.985\linewidth]{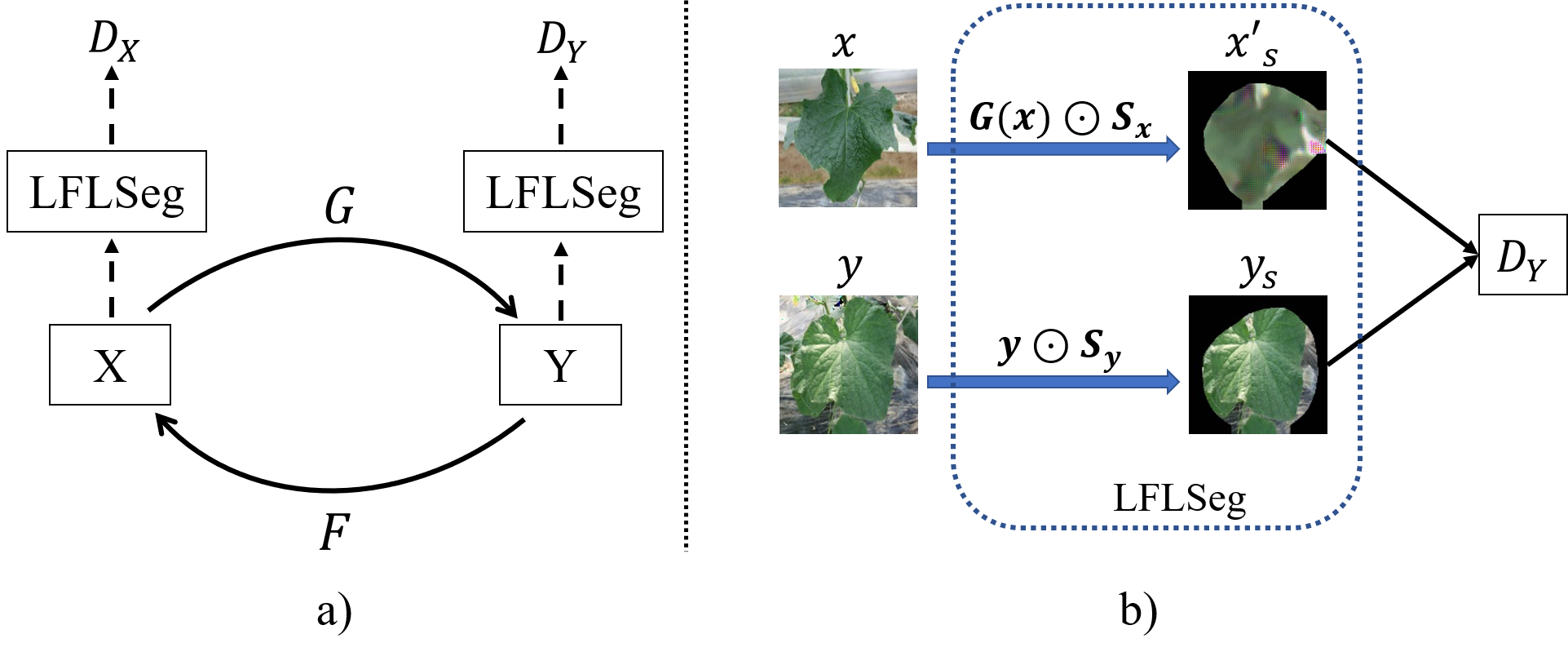}
\caption{
    \textbf{a)} Overview of the proposed LeafGAN scheme; \textbf{b)} Dataflow when transforming the sample $x \in X$ to the domain $Y$. Note that the dataflow from domain $Y$ to $X$ is the reverse of that from $X$ to $Y$. We use the same LFLSeg network in both transformations.
}
\label{fig:fig_2}
\end{figure}
LeafGAN is an image generation network that is specially designed to mitigate the serious overfitting problem in image-based plant diagnosis tasks via the effective generation of high-quality and widely varying pseudo training images. 
LeafGAN is built on CycleGAN and our proposed label-free leaf segmentation module (LFLSeg) to guide the network in transforming the relevant regions (i.e., leaf areas) while preserving the backgrounds. 
\figref{fig:fig_1} shows the limitations of the vanilla CycleGAN compared to LeafGAN; while CycleGAN transforms the entire image along with the background, LeafGAN focuses only on the leaf regions, resulting in natural and convincing generated images. 

Similar to CycleGAN, LeafGAN has two mapping functions  $G:X{\rightarrow}Y$ and $F:Y{\rightarrow}X$ corresponding to two data domains $X$ and $Y$. The training of $G$ requires a discriminator $D_Y$ to discriminate the generated image $G(x)$ from the real samples $y_i \in Y$. 
The mapping $F$ and the corresponding discriminator $D_X$, which discriminates the generated image $F(y)$ from the real samples $x_i \in X$, are also trained simultaneously. 
We assume here that $X$ and $Y$ are the sets of healthy and arbitrary target disease images, respectively. 

\figref{fig:fig_2}a shows an overview of the framework for LeafGAN. 
For the transformation $X{\rightarrow}Y$ (\figref{fig:fig_2}b), the proposed LFLSeg module first produces two binary masking images $S_x$ and $S_y$, which represent the leaf areas from input images $x \in X$ and $y \in Y$, respectively, where $S_x=\text{LFLSeg}(x)$ and $S_y=\text{LFLSeg}(y)$. 
After generating the image $x'=G(x)$, we obtain the masked leaf images $x'_s=S_x{\odot}x'$ and $y_s=S_y{\odot}y$, where $\odot$ denotes the element-wise product. 
These images $x'_s$ and $y_s$ are then fed into the discriminator $D_Y$ rather than feeding $x'$ and $y$. 
In this way, the discriminator is guided to discriminate only in terms of the leaf areas, instead of the backgrounds. 
Consequently, due to the adversarial training scheme for the generative adversarial networks \cite{goodfellow2014generative}, the generator $G$ is also forced to minimize its losses by paying attention to the leaf regions when generating (i.e., transforming) the images. 

Note that the dataflow for the transformation $Y{\rightarrow}X$ is the reverse of that for $X{\rightarrow}Y$, since they are symmetric. 
\begin{figure}[!t]
\centering
\includegraphics[width=0.98\linewidth]{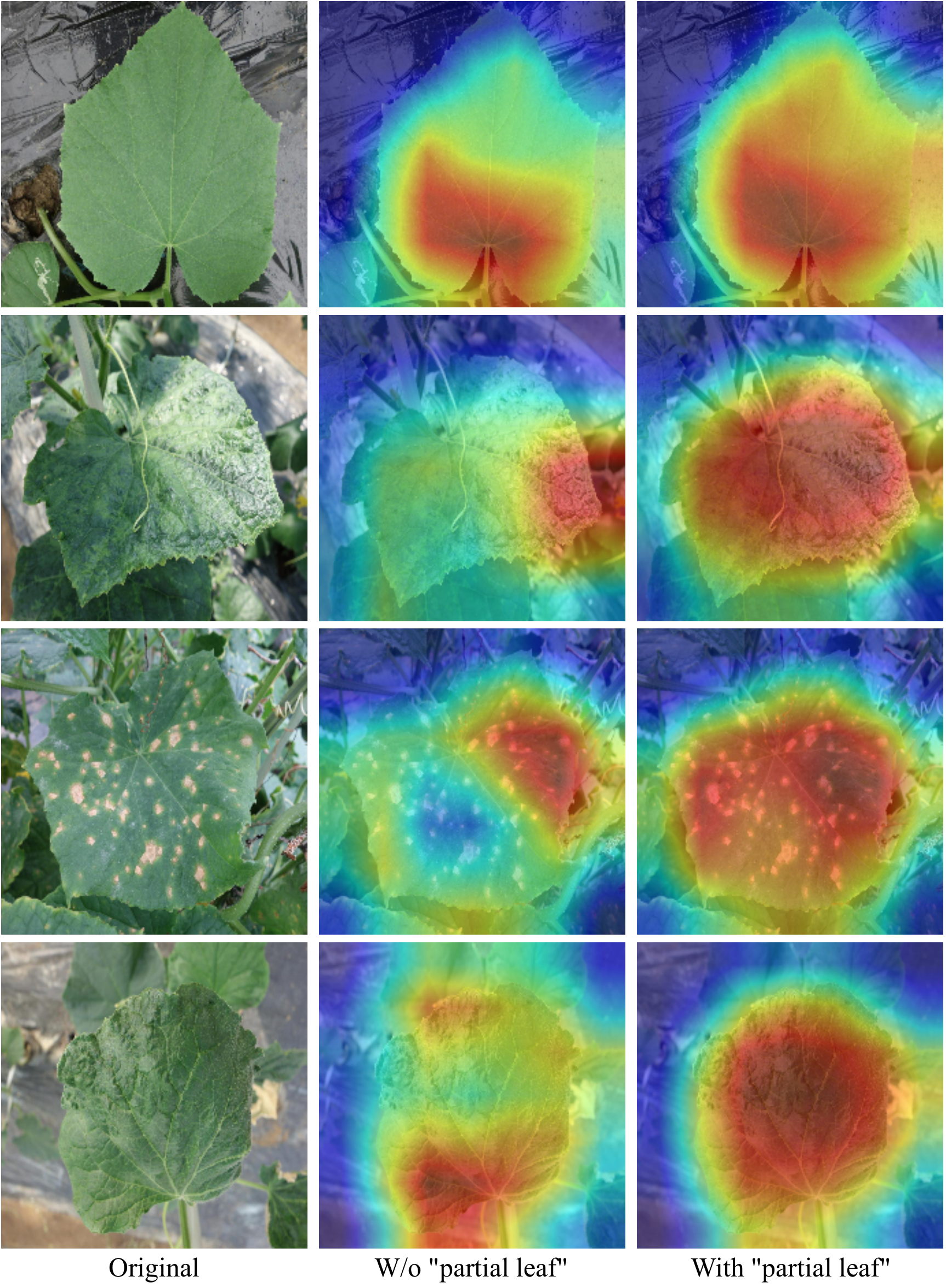}
\caption{
    The heatmaps comparison between LFLSeg models trained with and without the \protect\say{partial leaf} images. The warmer color region, the more it contributes to the final decision for a class (i.e., \protect\say{full leaf} in this case).
}
\label{fig:fig_3}
\end{figure}
\subsection{Label-free leaf segmentation module (LFLSeg)}
In practice, the segmentation of in-field leaf images using conventional techniques such as thresholding, clustering, edge detection, etc. is inefficient due to the complex appearance of the leaf and the diversity of the backgrounds as well as lighting conditions. 
A better option involves using the power of modern deep learning-based supervised \cite{shelhamer2017fully, ronneberger2015u, zhao2017pyramid, zhou2019semantic} or weakly supervised segmentation techniques \cite{oquab2015object, singh2017hide, li2018tell, lee2019ficklenet}. 
However, the former approach usually requires pixel-level annotation datasets in order to get a reliable result, and is therefore labor-intensive. 
As mentioned previously, our AOP model achieved an F1-score of 98.1\% for cucumber leaf segmentation, although this score was established using 8,000 masked images for training \cite{saikawa2019aop}. 
The latter approach extracts the segmentation information from feature maps produced by a deep network trained for image classification. 
Although the advantage of these weakly supervised models can be trained without extra labeling data, the models are often complex and require a lot of implementation. 

In this work, we propose a simple but effective weakly supervised label-free leaf segmentation module (LFLSeg) that helps the classification model to learn the dense and interior leaf regions implicitly. 
From an architecture point of view, the backbone of LFLSeg is a simple CNN, and is designed to discriminate between \say{full leaf}, \say{partial leaf}, and \say{non-leaf} objects. 
Specifically, \say{full leaf} objects are images that contain a single full leaf, while “partial leaf” objects are images that contain part of a \say{full leaf}, and \say{non-leaf} objects do not contain any part of a leaf. 

The segmented leaf region is obtained using a heatmap with respect to the “full leaf” class by applying the Grad-CAM \cite{selvaraju2017grad} technique. 
This heatmap is a probability map representing the contribution of each pixel to the final decision of the \say{full leaf} class, and thus can be used as a binary mask after thresholding with a specific threshold value $\delta$. 

The key idea underlying LFLSeg is the introduction of the \say{partial leaf} class for training. 
As mentioned in \cite{li2018tell}, a heatmap of a classifier that is trained to discriminate between an object and its background will only cover small and most discriminative regions of the object of interest. 
Hence, if we train our LFLSeg only to classify \say{full leaf} and \say{non-leaf} objects, the network will not be able to cover the \say{full leaf} area. 
The introduction of a \say{partial leaf} class leverages the model to seek a larger leaf-shaped region in order to classify the \say{full leaf} image correctly. 

\figref{fig:fig_3} shows a comparison of the heatmaps between classification models with and without the \say{partial leaf} training data. 
The warmer the color of a region, the more it contributes to the final decision for the \say{full leaf} class. 
These heatmaps show that our network, which was trained with the strategy described above, is able to focus on the whole shape of \say{full leaf} images, while the other model without the \say{partial leaf} class (i.e., which only classifies \say{full leaf} or \say{non-leaf} images) focuses on small, scattered leaf regions. 
\subsection{Loss functions for LeafGAN}
We design the loss functions for LeafGAN with reference to CycleGAN. 
The adversarial losses for the two mapping functions $G:X{\rightarrow}Y$ and $F:Y{\rightarrow}X$ are expressed as:
\begin{multline}
\mathcal{L}_\mathrm{adv}(G,D_Y)=\mathbb{E}_{y{\sim}p_\mathrm{data}(y)}[(D_Y(y_s)-1)^2]+\\
\mathbb{E}_{x{\sim}p_\mathrm{data}(x)}[(D_Y(x'_s))^2].
\end{multline}
Note again that $y_s=S_y{\odot}y$ is the masked version of the image $y \in Y$, where $S_y=\text{LFLSeg}(y)$ is the masking which represents the leaf area after feeding image $y$ to the LFLSeg module. 
Likewise, the adversarial loss $\mathcal{L}_\mathrm{adv}(F,D_X)$ for the mapping $F:Y{\rightarrow}X$ is defined as follows:
\begin{multline}
\mathcal{L}_\mathrm{adv}(F,D_X)=\mathbb{E}_{x{\sim}p_\mathrm{data}(x)}[(D_X(x_s)-1)^2]+\\
\mathbb{E}_{y{\sim}p_\mathrm{data}(y)}[(D_X(y'_s))^2].
\end{multline}
Please note that we use the same LFLSeg to segment the inputs from both domains $X$ and $Y$.
The cycle consistency loss is as follows:
\begin{multline}
\mathcal{L}_\mathrm{cyc}(G,F)=\mathbb{E}_{x{\sim}p_\mathrm{data}(x)}[|F(G(x))-x|_1]+\\
\mathbb{E}_{y{\sim}p_\mathrm{data}(y)}[|G(F(y))-y|_1].
\end{multline}

Since the purpose of our study is to enrich the backgrounds of images of diseased leaves, we need to prevent generating similar backgrounds as the images in the target domain and keep the generated backgrounds as close to the original input images as possible. 
To meet this requirement, we introduce a new loss term called \emph{background similarity loss} ($\mathcal{L}_\mathrm{bs}$). 
The objective of $\mathcal{L}_\mathrm{bs}$ is to minimize the $L_1$ distance between the background of the generated image and the original source image. 
The background can be easily obtained by calculating the element-wise product between the inverted version of the mask image $S$ (i.e., $1-S$) and the input leaf image. 
Therefore, 
\begin{multline}
\mathcal{L}_\mathrm{bs}(G,F)=\mathbb{E}_{x{\sim}p_\mathrm{data}(x)}[|(1-S_x){\odot}(G(x)-x)|_1 ]+\\
\mathbb{E}_{y{\sim}p_\mathrm{data}(y)}[|(1-S_y){\odot}(F(y)-y)|_1].
\end{multline}
Our final objective function is:
\begin{multline}
\mathcal{L}(G,F,D_X,D_X)=\mathcal{L}_\mathrm{adv}(G,D_Y)+\mathcal{L}_\mathrm{adv}(F,D_X)+\\
{\lambda}[\mathcal{L}_\mathrm{cyc}(G,F)+\mathcal{L}_\mathrm{bs}(G,F)],
\end{multline}
where $\lambda$ is a coefficient that controls the balance of different loss terms. 

\section{Experiments}
\begin{table}[t!]
\centering
\caption{Details of cucumber datasets (Datasets A and B)}
\label{tab:table_1}
\resizebox{0.47\textwidth}{!}{%
\begin{tabular}{|c|c|c|c|}
\hline
\multirow{2}{*}{\textbf{Class}} & \multicolumn{2}{c|}{\textbf{Dataset A}} & \textbf{Dataset B} \\ \cline{2-4} 
                                & \textbf{Training} & \textbf{Validation} & \textbf{Testing}   \\ \hline
Healthy (H)        & 4,000  & 717   & 1,046 \\ \hline
MYSV (H)           & 4,000  & 745   & 2,034 \\ \hline
Brown Spot (B)     & 2,000  & 784   & 1,220 \\ \hline
Powdery Mildew (P) & 2,000  & 796   & 89    \\ \hline
Total              & 12,000 & 3,042 & 4,389 \\ \hline
\end{tabular}%
}
\end{table}
\subsection{Cucumber diseases dataset}
In this work, we train our LeafGAN models to generate new images of cucumber disease. 
We collected cucumber leaf images from multiple locations in Japan, taken during the period 2015–2019. 
Each image contains a single cucumber leaf, roughly in the center and against various backgrounds. 
These images are of healthy (H) leaves or leaves infected with one of three diseases: Melon yellow spot virus (MYSV) (M), brown spot (B), or powdery mildew (P). 
\tabref{tab:table_1} summarizes the datasets used in our study. We divided these images into Datasets A and B. 
Images in these datasets were exclusive, and were taken on different farms. 
Dataset A was used for training and validation, and Dataset B was used to test performance. 
Note that the appearance of the images in those two sets varied due to the differences in the circumstances (e.g., photographic conditions and background) in which they were taken.
\subsection{Training the LFLSeg module}
We used the fine-tuned ResNet-101 model \cite{he2016deep} as the backbone of LFLSeg, and replaced the last layer of the network with a three-node layer. 
Using a deeper model (i.e., ResNet-152) yielded slightly better results, but we decided to use the ResNet-101 model for cost reasons. 
To train the LFLSeg module, we built datasets corresponding to the \say{full leaf}, \say{partial leaf}, and \say{non-leaf} classes. 
For the \say{full leaf} class, we used all 12,000 single leaf training images from Dataset A. 
During training of the network, we used a rotation with a step increment of 90 degrees and horizontal and vertical flips for data augmentation, giving a resulting dataset that was six times larger than the original one (i.e., 72,000 images). 

For the \say{partial leaf} class, we randomly selected 8,000 images from Dataset A (the training set) and divided each image into nine equally overlapping patches (i.e., 72,000 images). 
Given a \say{full leaf} image of size N$\times$N, we used a sliding window with size N/2$\times$N/2 to crop a training sample for \say{partial leaf} class with a step size of N/4$\times$N/4 from both the vertical and horizontal directions. 
In our preliminary experiments, we found that this setting showed the best performance. 

For the \say{non-leaf} class, 72,000 images were collected randomly from the ImageNet dataset \cite{deng2009imagenet}. 
In total, the training data for LFLSeg module consisted of 216,000 images. 
The dataset was divided randomly, with 70\% allocated as the training set and 30\% as the testing set.
Our LFLSeg module was fine-tuned using momentum optimization \cite{qian1999momentum} with a mini-batch size of 128. The training process was terminated after 30 epochs.
\begin{figure}[!t]
\centering
\includegraphics[width=0.98\linewidth]{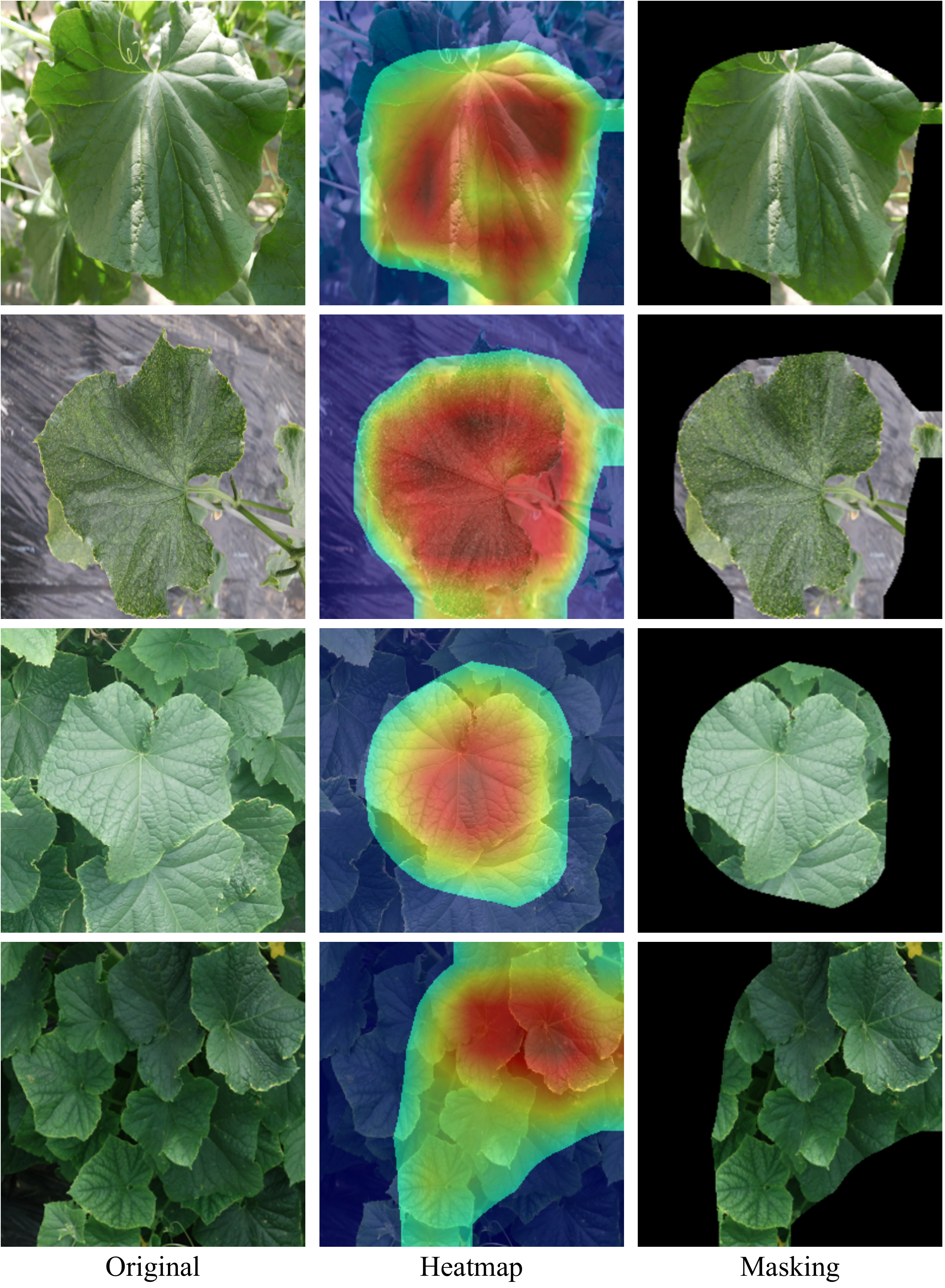}
\caption{
    Leaf segmentation results of the LFLSeg module. The heatmaps from our network can be used as the useful segmentation masks without the need of pixel-label data.
}
\label{fig:fig_4}
\end{figure}
\subsection{Training the disease translation models}
We used LeafGAN to build three types of healthy$\leftrightarrow$diseased translation models: (i) healthy$\leftrightarrow$MYSV (H$\leftrightarrow$M); (ii) healthy$\leftrightarrow$brownspot (H$\leftrightarrow$B); and (iii) healthy$\leftrightarrow$powderymildew (H$\leftrightarrow$P). 
For comparison purposes, we also built three more corresponding disease translation models using CycleGAN. 

Since there were only 2,000 training samples for each of the brown spot (B) and powdery mildew (P) classes, we randomly selected 2,000 images from 4,000 images of healthy leaves (H) to train the (H$\leftrightarrow$B) and (H$\leftrightarrow$P) models (i.e., we used a total of 2,000 healthy images in this case). 
Note that we only used one-way translation from healthy$\rightarrow$diseased at test time, since our target was to generate more data for diseased leaves. 

We applied the same parameters as described in \cite{zhu2017unpaired} to train both the CycleGAN and LeafGAN models. 
For the LeafGAN model, we set the segmentation threshold value for LFLSeg to ${\delta}=0.35$. 
Training of both the LeafGAN and CycleGAN models was terminated after 200 epochs. 
Please refer to the CycleGAN article for more details of the training process. 

At test time, we generated new three types of disease images from healthy images from the validation set of Dataset A (i.e., 717 images for each disease type in our experiments). 
These images were then used as augmented data for further training of the disease classifiers. 

\subsection{Training the disease classification models}
To carry out a qualitative evaluation of the effectiveness of LeafGAN in terms of improving the generality of disease diagnosis performance on an unseen dataset, we trained the disease diagnosis models with and without images newly generated by LeafGAN, and compared the performance in each case. 
Specifically, we trained the following classifiers:
\begin{itemize}
  \item[-] The first classifier was trained using only the training images from Dataset A. 
  We refer to this as our baseline model.
  \item[-] The second classifier was based on the above baseline model but was trained with additional disease images generated by the CycleGAN models. 
  We refer to this as baseline+CycleGAN.
  \item[-] The third classifier was similar to the second classifier, but was trained with additional disease images generated by the LeafGAN models. 
  We refer to this as the baseline+LeafGAN. Note that this is the proposed model. 
\end{itemize}

All classifiers were fine-tuned from the pretrained ResNet-101 model, and we applied horizontal and vertical flip augmentation on the fly during training. 
The SGD momentum optimizer with a minibatch size of 128 was used to train these models. The training process was terminated after 30 epochs.

\section{Results}
\begin{table}[!t]
\centering
\caption{Performance comparison of the three classifiers in disease diagnosis on the unseen Dataset B}
\label{tab:table_2}
\resizebox{0.47\textwidth}{!}{%
\begin{tabular}{|c|c|c|c|c|}
\hline
\textbf{Class} &
  \textbf{\# of test images} &
  \textbf{\begin{tabular}[c]{@{}c@{}}Baseline\\ (\%)\end{tabular}} &
  \textbf{\begin{tabular}[c]{@{}c@{}}Baseline+\\ CycleGAN\\ (\%)\end{tabular}} &
  \textbf{\begin{tabular}[c]{@{}c@{}}Baseline+\\ LeafGAN\\ (\%)\end{tabular}} \\ \hline
Healthy (H)        & 1,046 & \textbf{85.1} & 84.7 & 84.6 \\ \hline
MYSV (M)           & 2,034 & 75.4 & 76.0 & \textbf{83.3} \\ \hline
Brown Spot (B)     & 1,220 & 62.8 & 65.4 & \textbf{75.9} \\ \hline
Powdery Mildew (P) & 89    & 61.8 & 61.8 & \textbf{70.8} \\ \hline
Average            &       & 71.3 & 72.0 & \textbf{78.7} \\ \hline
\end{tabular}%
}
\end{table}
\subsection{Segmentation performance of LFLSeg}
Our LFLseg module achieved an accuracy of 99.8\% in classifying the three classes (\say{full leaf}, \say{partial leaf}, \say{non-leaf}) on the validation set from Dataset A. 
\figref{fig:fig_4} shows several examples of leaf segmentation using our proposed LFLSeg module, with heatmaps for the \say{full leaf} class and their corresponding segmented results. 

We confirmed that LFLSeg works well on different in-field images with complex backgrounds. 
However, when the images contain multiple and overlapping leaves, the LFLSeg fails to correctly segment the leaf area (\figref{fig:fig_4} last row). 
Despite this fact, we do not expect the input which contains multiple leaves to be the case since we assume the input of the disease classifier is a single leaf image in this study. 

We also compared the segmentation performance of our LFLSeg module with the previous AOP network \cite{saikawa2019aop} using 1,000 full leaf images. 
Our module achieved an F1-score of 83.9\% while the AOP network achieved 98.1\%. 
Even though LFLSeg shows poorer performance than the AOP network with pixel-level labeled training images, our network which requires no masking training data still achieved a reliable result that was sufficient for our task. 
\begin{figure*}[t!]
\centering
\includegraphics[width=0.95\linewidth]{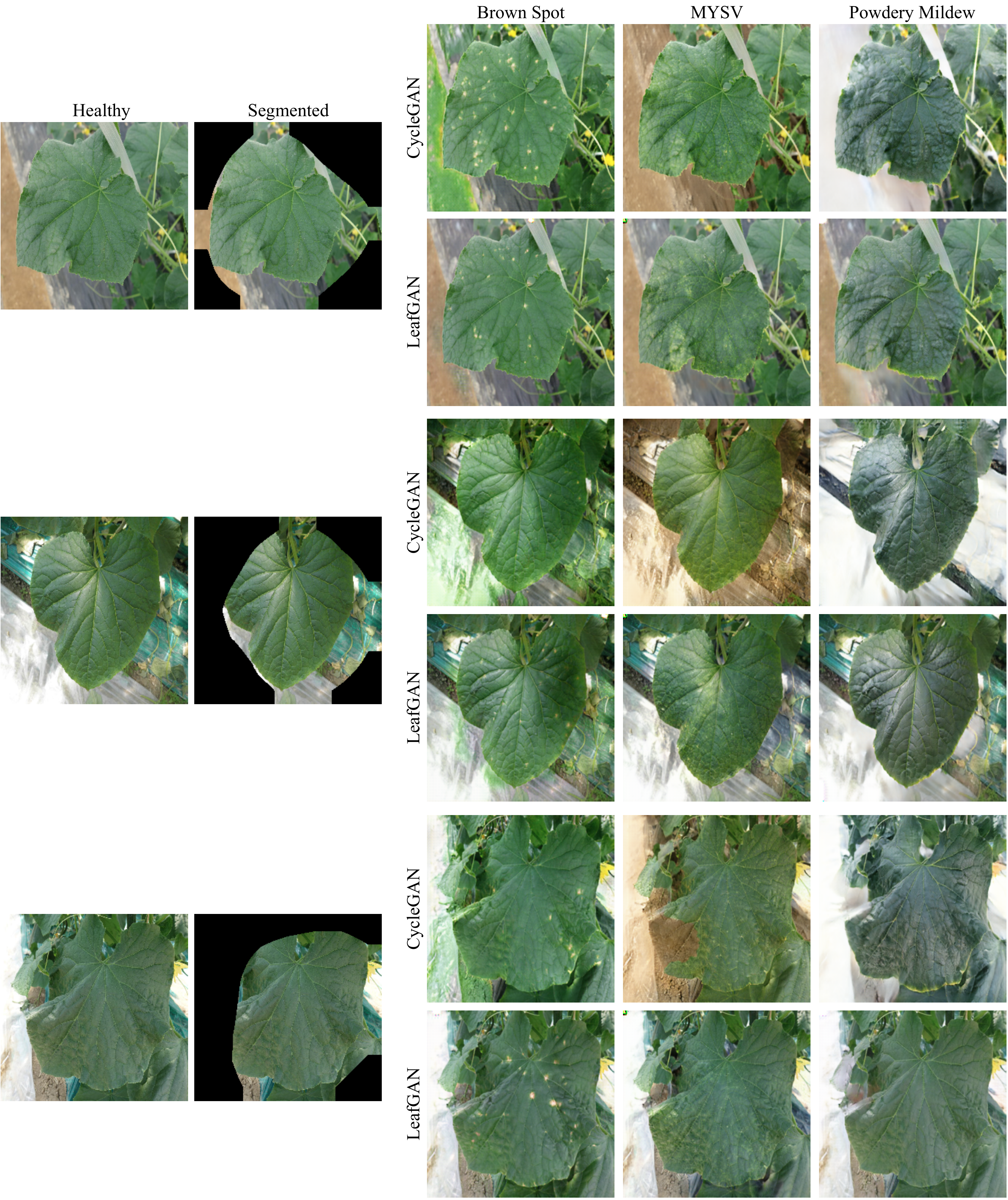}
\caption{
    Comparison of the images generated by CycleGAN and LeafGAN. The segmented leaves are the outputs of our LFLSeg module. LeafGAN preserves the background from the original, meaning that the generated images are more realistic than those of CycleGAN.
}
\label{fig:fig_5}
\end{figure*}
\subsection{Results from disease translation models}
Examples of diseased images generated by the CycleGAN and LeafGAN models are shown in \figref{fig:fig_5}. 
Without the explicit attention mechanism, CycleGAN tends to transform the whole area of the image, including the background, giving implausible results. 
In contrast, our LeafGAN models learn to pay attention to the leaf regions rather than the backgrounds, which gives more realistic disease images. 

\subsection{Improving the generality of disease diagnosis systems}
The baseline, baseline+CycleGAN, and baseline+LeafGAN models obtained average accuracies of 97.2\%, 97.7\%, and 97.9\%, respectively, on the validation images from Dataset A. 
We then used the trained classifiers to test Dataset B. 
\tabref{tab:table_2} presents a comparison of the diagnostic performance of the three classifiers on the unseen Dataset B. 

We can see that there is a large gap between the average accuracy for the Dataset A validation set and Dataset B, since the two sets are completely different. 
Although the baseline model was trained on 2,000 images per class, the average diagnostic performance only reached 71.3\%. 
The LeafGAN models helped to boost the performance, and achieved the best result of the three classifiers with an average accuracy of classification of 78.7\%. 

\section{Discussion}
    In this study, we investigated the effectiveness of using image-to-image translation models as a data augmentation tool to improve the performance of an automated diagnosis system for cucumber plant disease. 
In this experiment, the baseline model was overfitted to its training dataset and did not generalize well to the unseen samples. 
We also observed a large performance gap on training and testing datasets as noted in former studies \cite{mohanty2016,saikawa2019aop, suwa2019comparable}. 
The visual results in \figref{fig:fig_5} demonstrate that our LeafGAN model can generate more persuasive and realistic images than the original CycleGAN. 
Since CycleGAN learns to transform the whole content of the training images, the backgrounds of the generated results appear closer to the samples from the target domain. 
Specifically, the backgrounds of the healthy images are transformed to be as close as possible to the images from the real disease datasets (see \figref{fig:fig_5}). 
\begin{figure}[!t]
\centering
\includegraphics[width=0.99\linewidth]{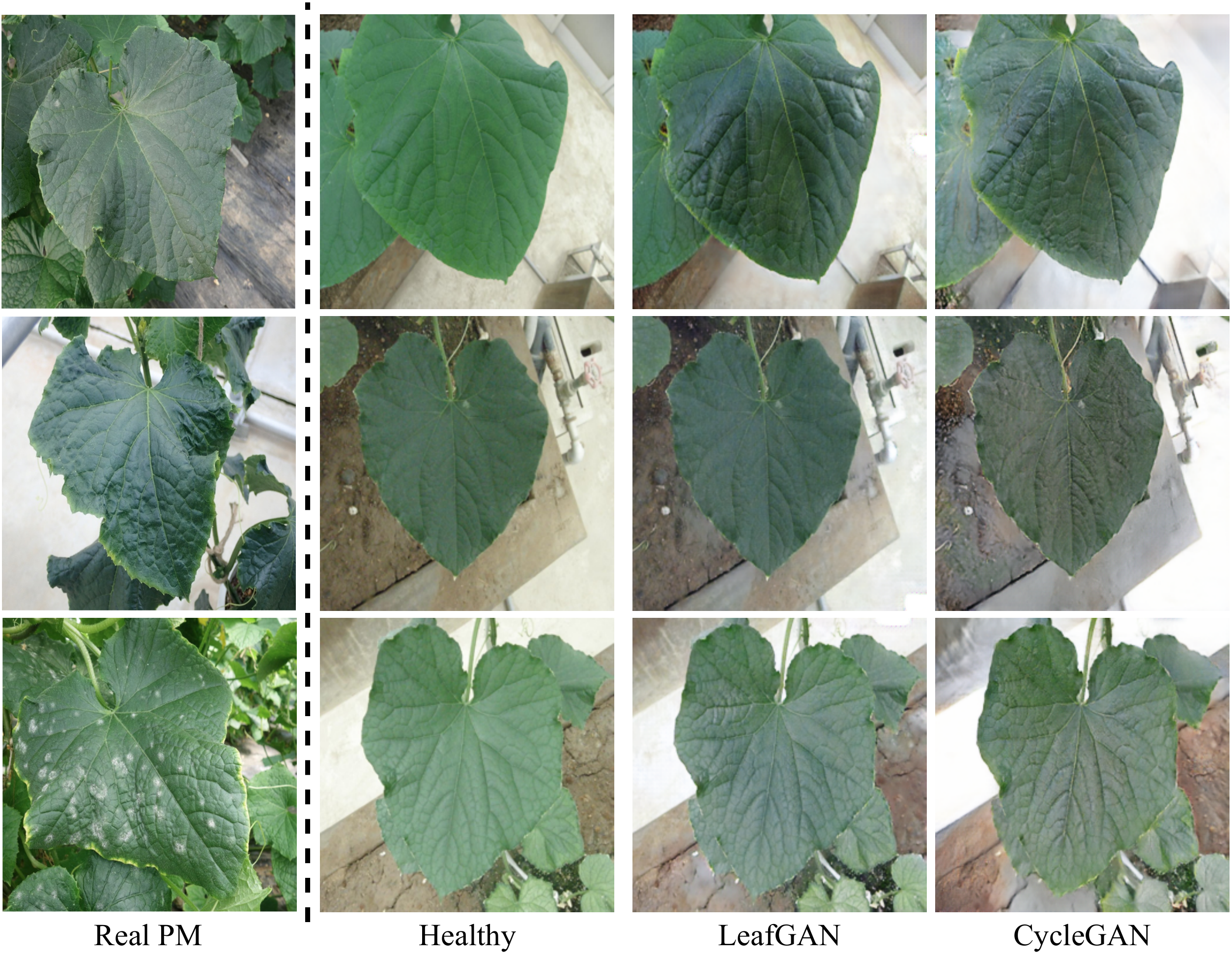}
\caption{
    Symptoms of the PM disease in different stages (left column) and the failure cases of the H$\rightarrow$P models from both LeafGAN and CycleGAN when translating from healthy images (second to last column).
}
\label{fig:fig_6}
\end{figure}

Using the proposed LFLSeg module, our LeafGAN is guided to focus on transforming only the leaf area, and can generate more compelling results. 
Although LFLSeg does not perfectly segment the leaf region, it is sufficiently effective to guide the LeafGAN models, thanks to the introduction of the \say{partial leaf} class. 
The results in \tabref{tab:table_2} show that using LeafGAN as a data augmentation tool can improve the diagnostic performance by 7.4\% on the unseen Dataset B. 
This is because the probability distribution of the generated images is significantly different from that of the original training data, due to the integration of images by the segmentation mask. 
The intrinsic variety of the training data therefore increases from a stochastic point of view. 
In addition, symptoms appear only in the relevant region, and we believe this is an advantage in boosting the classification performance. 
We believe that improving the performance of LFLSeg or combining LeafGAN with a sophisticated segmentation system such as AOP could improve the quality of the generated images. 

The results from the baseline+CycleGAN model showed that if we simply trained the disease classifier with the generated images from the CycleGAN models, which have no attention mechanism, the diagnostic performance improved only slightly (+0.7\%) compared to the baseline model. 
This is because the disease images generated by CycleGAN are intended to have a close probabilistic distribution to the training disease data, and thus the variety of training data is increased only a little. 
This is also discussed in the literature \cite{perez2017effectiveness, han2018infinite, shmelkov2018good, ma2018background}. 
From a visual assessment, it can be more intuitively seen that because CycleGAN tries to generate overall images that are probabilistically similar to the training disease images, the symptoms are often generated in the surrounding areas, meaning that the disease classifier may use background areas as the discriminative regions. 

Although our system achieved promising results, we still observed two remain limitations. 
\emph{First}, the proposed LFLSeg may incorrectly detect \say{partial leaf} as \say{full leaf} if the \say{partial leaf} image has a different shooting distance than images in our training dataset. 
Even though we rarely encounter this extreme case, applying data augmentation techniques such as random resize/scale is expected to increase the robustness of this module and thus, boosting the performance of our system for future usage. 
\emph{Second}, due to the complex characteristics of the training dataset, both LeafGAN and CycleGAN sometimes transformed the color rather than the disease symptom. 
\figref{fig:fig_6} shows the characteristics of the powdery mildew (PM) disease in different stages (left column) and the failure cases of the H$\rightarrow$P models from both LeafGAN and CycleGAN when translating from healthy images (second to last column). 
The PM dataset contained leaf images (left column) that were mostly in early and middle stages (first two images) with many of them are in dark blue color, while the later stage of PM disease (last image) is a typical case, but there is little in our dataset. 
Therefore, in several cases, the models generated images with a different color and few signs of PM symptoms. 
We believe that there is room for improvement in our system by addressing these practical problems, and we intend to investigate this in future work. 

\section{Conclusion}
    This paper proposes the LeafGAN method as an effective data augmentation tool for improving the robustness of an automated plant disease diagnosis system. 
With the ability of segmenting leaf areas and transforming a diverse range of backgrounds, LeafGAN demonstrates significant improvements in the quality of the generated images and boosting the overall disease diagnosis performance. 
We believe that our LeafGAN method is a reliable data augmentation tool and will make a significant impact on the field of automated crop disease diagnosis. 
\section*{Acknowledgment}
This research was partially supported by the Ministry of Education, Culture, Science and Technology of Japan (Grant in Aid for Fundamental research program (C), 17K8033, 2017-2020).

\nocite{*}
\footnotesize{
\bibliographystyle{IEEEtran}
\bibliography{reference}
}

\end{document}